\documentclass[Afour,sageh,times]{sagej}

\usepackage{moreverb,url}

\usepackage[colorlinks,bookmarksopen,bookmarksnumbered,citecolor=red,urlcolor=red]{hyperref}

\usepackage{graphicx}
\usepackage{amsmath}
\usepackage{amssymb}
\usepackage{algorithm}
\usepackage[noend]{algpseudocode}
\usepackage{tabularx}
\usepackage{subfig}
\usepackage{alphalph}

\usepackage{xcolor}
\newcommand\BibTeX{{\rmfamily B\kern-.05em \textsc{i\kern-.025em b}\kern-.08em
T\kern-.1667em\lower.7ex\hbox{E}\kern-.125emX}}

\def\volumeyear{2015}

\begin{document}

\runninghead{Broad and Argall}

\title{Geometry-Based Region Proposals for Real-Time Robot Detection of Tabletop Objects}

\author{Alexander Broad\affilnum{1,2} and Brenna Argall\affilnum{1,2}}

\affiliation{\affilnum{1}Northwestern University, Chicago, IL
\affilnum{2}Rehabilitation Institute of Chicago, Chicago, IL
}

\corrauth{Alexander Broad, Department of Electrical Engineering and Computer Science,
Northwestern University
633 Clark St, 
Evanston, IL 60208}

\email{alex.broad@u.northwestern.edu}

\begin{abstract}
We present a novel object detection pipeline for localization and recognition in three dimensional environments.  Our approach makes use of an RGB-D sensor and combines state-of-the-art techniques from the robotics and computer vision communities to create a robust, real-time detection system.  We focus specifically on solving the object detection problem for tabletop scenes, a common environment for assistive manipulators.  Our detection pipeline locates objects in a point cloud representation of the scene. These clusters are subsequently used to compute a bounding box around each object in the RGB space.  Each defined patch is then fed into a Convolutional Neural Network (CNN) for object recognition.  We also demonstrate that our region proposal method can be used to develop novel datasets that are both large and diverse enough to train deep learning models, and easy enough to collect that end-users can develop their own datasets.  Lastly, we validate the resulting system through an extensive analysis of the accuracy and run-time of the full pipeline. 
\end{abstract}


\maketitle

\section{Introduction}

As the field of robotics advances, and personal robots that assist users in their home and work environments become more prevalent, it will be necessary to extend a robot's autonomy to include more advanced cognitive reasoning and improve abilities in highly dynamic environments.  To do so, the robot will need to have knowledge of many of the same physical attributes of the world that a human does.  One such important aspect is the ability to recognize and localize objects in common environments.  Through this knowledge, a robot can make informed decisions in achieving tasks like intelligently searching for an object in a novel environment, cleaning a room or retrieving an object for a human partner.  

The problem of object detection is not unique to robotics.  In computer vision it is used to solve problems like automatic caption generation as demonstrated by \cite{karpathy2015deep} and automatic tagging of shared social pictures as described in \cite{schroff2015facenet}.  However, it is often the case that techniques used in the two communities are distinct from one another.  One reason for this is that the desired and available sensor information is frequently different --- in computer vision, systems are usually limited to the RGB space while solving problems in robotics generally requires depth as an additional, or primary, modality.  The requirement of depth information often necessitates an additional sensor (with a few notable exceptions such as \cite{saxena2009make} and \cite{mur2015orb}), which in turn requires a potentially difficult calibration and sensor fusion problem.  For this reason, we frequently see methodologies in the two communities that parallel each other in purpose, such as object recognition, but are divergent in technique.  However, due to the rise of RGB-D cameras like the Microsoft Kinect (\cite{zhang2012microsoft}), robotics researchers have access to sensors that provide both color and depth information in a single device.  These cameras can be easily calibrated (up to a level of tolerance) and aligned through a single transformation defined by the static configuration of the two integrated sensors.

In this paper, we propose a novel method for object detection that makes use of both the depth and color modalities of RGB-D sensors to recognize and localize objects in real-time.  We focus specifically on tabletop environments as many domestic manipulation tasks take place in this type of configuration.  Our approach differs from previous proposed techniques in that it solves the localization and recognition tasks independently --- the former through an exploitation of the geometry of the scene, and the latter with state-of-the-art deep learning methods.  To the best of our knowledge, this work is the first to combine these ideas.  Our method achieves high accuracy in both the position and categorization of the detected objects by using a confluence of ideas from the computer vision and robotics communities.

We begin by discussing related work in Section~\ref{sec:related} and then present our approach in detail in Section~\ref{sec:approach}.  We also discuss how our region proposal method can be used for data acquisition and developing novel datasets in Section~\ref{sec:create_dataset}.  Finally, we describe an experimental validation of our system in Section~\ref{sec:exp} and the results in Section~\ref{sec:results}.  We then discuss the success of our approach and future directions in Section~\ref{sec:disc} and conclude in Section~\ref{sec:conclusion}.


\section{Related Work}
\label{sec:related}

From a computational perspective, object detection is a two part problem:  (1) Where is the object? and (2) What is the object? The long-standing baseline approach in computer vision is known as a sliding window.  In this technique, each patch of size $(m,n)$, from an image of size $(M,N)$, is fed through an object recognition model.  The concept is that, while this approach may be inefficient, it maximizes recall by ensuring not to skip any possible object locations.  To account for scale the same process is repeated over an image pyramid. 
 
More recently, as real-time and interactive systems have become more popular (\cite{chen2005time}) there has been an increased focus on the efficiency of object detection systems.  The most common way to improve the run-time of the system is to intelligently reduce the number of candidate regions that are run through the object recognition model.  New approaches focus on novel techniques and methods for producing \textit{region proposals}.  For example, \cite{girshick2015fast} uses multi-layer segmentation to produce region proposals at different positions and scales, \cite{szegedy2013deep} train a neural network to predict segmentation masks and \cite{zitnick2014edge} use edge detections.  \cite{hosang2014good} provide a comprehensive comparison of region proposal techniques in the computer vision community.  By reducing the number of candidate regions, one is able to perform a more efficient search through position, scale and orientation.  

These techniques greatly reduce the computational burden when compared to an exhaustive sliding window approach, however they often still require expensive systems and GPUs to train and run.  For example, Fast R-CNN, as proposed by \cite{girshick2015fast} has proven very successful, yet when using this approach on a 640$\times$480 image with a Core i7 laptop with a mid-tier GPU (nVidia GeForce 860M), the full pipeline takes about 0.75 seconds per image.  One reason for this is that the segmentation algorithm produces between 1k and 10k proposals per image depending on the \textit{quality mode} parameter.  To increase the speed of this system, \cite{ren2015faster} extend Girshick's method to a model called Faster R-CNN, which uses a separate neural network to produce object proposals, decreasing the run-time to about 0.2 seconds per image.  Of note, these approaches necessitate significantly more data as the training process requires labeled bounding boxes and an extra background class to reject false positives.  Additionally, localizing an object in the 2D plane does not fully solve the problem for robotics applications where localizing the object in three dimensions is equally as important as correctly recognizing the object.  Lastly, there is also evidence from \cite{chen20153d} that image-based segmentation approaches are not equally effective on all datasets.  

There also is related work from the robotics community in 3D object detection.  Early approaches are similar to pre-deep learning methods in the vision community; namely, they focus on developing hand-crafted features in the point cloud space.  Examples include local features such as the histogram-based Fast Point Feature Histogram of \cite{rusu2009fast} and the Signature of Histogram of Orientations of \cite{tombari2010unique}, as well as global features such as the Viewpoint Feature Histogram of \cite{rusu2010fast} which also takes the viewpoint into account.  \cite{tang2012textured} describe a segmentation approach similar to our own, however the recognition is again done in the point cloud space by comparing features to learned object models.  The approaches mentioned here work relatively well, however they rely on hand-crafted features and no single approach has found the type of success or wide-spread adoption of Convolutional Neural Networks in the image space (\cite{lecun2015deep}).  

More closely related to our own work, researchers have begun to look at other methods for combining the RGB and depth modalities when solving the object detection problem.  \cite{song2014sliding} describe a three dimensional version of the sliding window approach in which they fit 3D regions to learned CAD-based object models.  \cite{dahan2012combining} describe a method for computing region proposals by segmenting the input image using information from both the color and depth channels.  \cite{couprie2013indoor} and \cite{gupta14rcnndepth} similarly describe methods for including depth during segmentation and they also augment the standard vision approach by including the depth map as another input channel in training the CNN model.  The main difference between these works and our own is that we explicitly generate region proposals in the point-cloud space based on geometric constraints, which produces significantly fewer candidate regions.  \cite{pillai_rss15} present a robotic recognition system that incorporates multi-view object proposals and efficient feature encoding methods to solve a similar problem.  In particular the researchers develop a SLAM-aware system that incorporates a detection model to improve robotic object recognition.  However, again, this work is distinct from our own as it performs recognition only in the point cloud space.

Lastly, \cite{chen20153d} describe a 3D object proposal method that is particularly focused on autonomous driving (e.g. detected objects include cars, pedestrians and cyclists).  In this work, researchers similarly use known geometric features to reduce possible candidate regions and then propose an energy minimization formulation to compute region proposals.  This approach places a greater emphasis on finding the \textit{best} bounding box for each object, which requires additional prior information such as object size priors, point densities and free space information.  As we perform localization in the point cloud space, the fit of the bounding box to a hand-labeled source is not nearly as important.  While tight bounding boxes may be important in applications like self driving cars (as this information may be necessary for both high level planning and low-level dynamics considerations), the increased fidelity requires additional computation and therefore the system runs at an average of 1.2 seconds per image (with N=2000 proposals) at runtime.  We instead focus on domestic environments (like \cite{rusu2009close} and \cite{stuckler2013efficient}), which allows us to retain the desired recognition accuracy at significantly improved speed by relaxing the requirements on our region proposal to \textit{any} bounding box suitable for object recognition. 

\section{Our Approach}
\label{sec:approach}

\begin{figure*}[t]
    \centering
    \includegraphics[width=\textwidth]{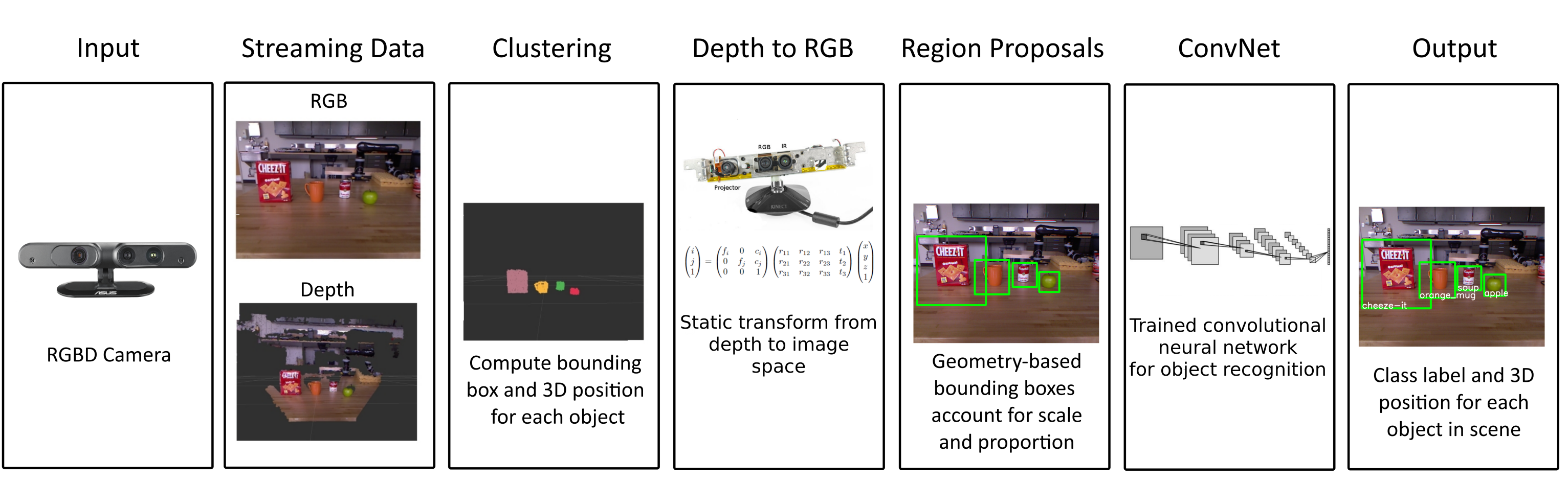}
    \caption{High-level step-through of our object detection pipeline.  From left to right: We use an RGB-D camera to capture color and depth information about our scene.  We then exploit known geometric properties to compute a bounding box and 3D position for each visible object.  The bounding boxes are translated to RGB space using the static transform defined by the position of the two sensors in the camera.  These patches are fed through a trained CNN for object recognition.  The output is a class label and 3D position for each object in the scene.}
    \label{fig:snp}
\end{figure*}  

Our approach leverages both RGB and depth sensing modalities in a single object detection pipeline.  In this work we focus specifically on detection in a tabletop setting, a common environment for assistive manipulators and particularly useful to researchers.  We take inspiration from the computer vision community and develop a novel region proposal method, however, our technique is rooted in robotics perception and makes use of three dimensional point cloud data.  To do so, we exploit the geometry of the environment to produce a minimum set of region proposals described in Section~\ref{sec:a}.  We then translate our candidate regions from three dimensional bounding boxes into their two dimensional representation in the image plane, described in Section~\ref{sec:b}.  This image patch is then fed into a CNN for classification, described in Section~\ref{sec:c}. The complete approach is outline in Figure~\ref{fig:snp}.



\subsection{Object Localization}
\label{sec:a}


Our object localization method is detailed in Algorithm~\ref{algo:region}.  This algorithm simultaneously computes a bounding box, $\mathbb{B}$, and three dimensional position, $\mathbb{P}$, of each object in the scene.  The localization method capitalizes on known geometric properties of the table to reduce the computational burden and produce highly reproducible results.

The input to the algorithm is a point cloud, $C$, which we then downsample (Line 2) to ensure coverage and speed.  We downsample the input with a voxel filter which reduces the number of voxels necessary to represent the scene by replacing each set of $V$ voxels with a single voxel located at their centroid.  The downsampling parameter, $\alpha$, defines the fraction of voxels in our final representation compared to the input point cloud --- in our experiments this was set to 0.1.  An optional step to further reduce the computational burden is to also run the point cloud through pass-through filters parameterized by the geometry of the tabletop (Line 3).  These filters removes voxels in the scene that are outside the physical bounds of the table.  Our experimental results in Section~\ref{sec:results} use this step.  We can further remove any points belonging to the tabletop itself by using the random sample consensus  (RANSAC) method (Lines 4-5) to find the dominant plane, $T$, in the point cloud scene, $C$.  We can then filter these voxels from the remaining scene to remove any voxels belonging to the table.  A Euclidean clustering algorithm is run on the remaining points to discover continuous objects in the scene (Line 6).  We can then compute a bounding box $\mathbb{B}$ around each object in the set of clusters $o \in \mathbb{O}$ by finding the upper left, $U$, (Line 9) and lower right, $L$, (Line 10) corners of the cluster $o$.  We also compute the three dimensional position $\mathbb{P}$ of an object by computing its centroid (Line 12).

\begin{algorithm}
\caption{Geometric Region Proposal}\label{region_proposal}
\begin{algorithmic}[1]
\State $\textbf{Given }\textnormal{Point Cloud }C, \textit{optional:} \textnormal{ table dimensions}$
\State $C \gets downsample(C, \alpha)$
\State $\textit{optional: } C \gets passthrough(C, \textnormal{ table dimensions})$
\State $T_{inliers} \gets RANSAC(C)$
\State $T_{outliers} \gets C - T_{inliers}$
\State $\mathbb{O} \gets Cluster(T_{outliers})$
\State $\textbf{Init } \mathbb{B} \gets \emptyset, \mathbb{P} \gets \emptyset$
\For {$o \in \mathbb{O}$}
\State $U \gets (x_{min}, y_{max}, z_{max})$
\State $L \gets (x_{max}, y_{min}, z_{max})$
\State $\mathbb{B} \cup (U, L)$
\State $\mathbb{P} \cup centroid(o)$
\EndFor
\State $\textbf{return} \hspace{2mm} \mathbb{B}, \mathbb{P}$
\end{algorithmic}
\label{algo:region}
\end{algorithm}

Similar to other model-free segmentation approaches, a benefit of our method is that the region proposal algorithm does not rely on learning a model from a large dataset.  Instead, we make use of the geometry of the scene to develop candidate object locations.  Therefore, as only the recognition portion of the pipeline happens in the image-space, our model does not require training data that includes additional meta-data such as object bounding boxes.  This approach has the added benefit of significantly decreasing the number of region proposals when compared to other methods.  That is, for each object in the scene we propose only a \textit{single} region by using the physical properties of the object to account for both position and scale.  Our method is particularly well-suited for our problem domain as a vast number of manipulation objects are easily clustered due to their shape and size.  Additionally, even in cluttered environments, the depth dimensionality helps separate objects that otherwise look nearby in RGB space.  Another benefit to computing the region proposals in the depth modality is that the localization of the object is very accurate due to the resolution and precision of the RGB-D sensor (\cite{khoshelham2012accuracy}).  For example, in our experiments the table was one meter wide, indicating a maximum error of $\sim6mm$.\footnote{The error in depth measurements using an RGB-D sensors is calculated through triangulation.  The error increases with the distance squared as described in \cite{zhang2012microsoft}.} 



\subsection{Translation between Depth and RGB space}
\label{sec:b}

The next step in our object detection pipeline is to classify each proposal region.  We choose to perform the classification in the image space due to the demonstrated accuracy and expressivity of deep learning methods.  Therefore, we must translate the bounding box from the depth frame into the image frame.  The coordinates of the bounding boxes in these two modalities are not directly aligned due to a physical offset in the sensor, however we can compute the transformation between them as described in \cite{karan2015calibration}.  

To transform the bounding box in depth space to its representation in RGB space, we can begin by representing the RGB-D sensor as a pinhole camera.  Under this assumption, each point in the depth space $(x,y,z) \in \mathbb{R}^3$ and each point in the image space $(i,j) \in \mathbb{R}^2$ is mapped into its homogeneous coordinate definitions, $(x,y,z,1)$ and $(i,j,1)$ respectively.  We can then define a projective relationship between the two representations based on the intrinsic $(f, c)$ and extrinsic $(r, t)$ parameters of the camera as seen in Equation~\ref{eq:d_to_rgb}.
\begin{equation}
\begin{pmatrix} i \\ j \\ 1\end{pmatrix} = 
\begin{pmatrix} f_i & 0 & c_i \\ 0 & f_j & c_j \\ 0 & 0 & 1 \end{pmatrix} 
\begin{pmatrix} r_{11} & r_{12} & r_{13} & t_{1} \\ r_{21} & r_{22} & r_{23} & t_{2} \\ r_{31} & r_{32} & r_{33} & t_{3} \\\end{pmatrix} 
\begin{pmatrix} x \\ y \\ z \\ 1\end{pmatrix}
\label{eq:d_to_rgb}
\end{equation}

In this equation the first matrix represents the camera's intrinsic parameters and describes a transformation between the optical center of the camera and a given point in the image frame.  Specifically, $f_i, f_j$ represent the focal length in pixel space and $c_i, c_j$ represent the physical offset between the origins of each frame in pixel space.  The second matrix represents the camera's extrinsic parameters consisting of a rotation matrix, $R$, and the position of the origin in the world frame, $T$, which describes a transformation between the position and orientation of the depth- and RGB-cameras.  These are defined by the sensor hardware and are often both readable and tunable using the associated driver (the values used in our implementation are included in the open source code).  Through the use of Equation~\ref{eq:d_to_rgb}, we can therefore translate each bounding box in the point cloud to a bounding box in the image space.

Until this point, the bounding box we have computed tightly constrains each object in the scene, however, for the recognition portion of our pipeline it is useful to have a border around the object itself.  This is because most image based recognition networks are trained with patches that include a border around the object of interest.  For this reason, we slightly expand the bounding box associated with each object.  The size of the border can be tuned (in our work, we expanded the border by $40\%$), however the same parameters should be used during data collection and at runtime.

\subsection{Object Recognition}
\label{sec:c}


The final step in our object detection pipeline is recognition, which we solve using a convolutional neural network.  Each region proposal is extracted from the full image, scaled to the input size needed for our trained model and classified.  

The specific network architecture chosen for the classification portion of the pipeline is easily replaced and adjustable to stay in line with the state-of-the-art in deep learning.  In our experiments (see Section~\ref{sec:exp}), we evaluate a small model that we train from scratch as well as three state-of-the-art architectures initialized with weights learned on the ImageNet (\cite{krizhevsky2012imagenet}) dataset and finetuned on our dataset.  Importantly, using larger CNN models does not have a particularly large effect on the runtime of our system as we only evaluate the recognition model once per region proposal.  Since there are no widely circulated networks weights trained on a dataset that encompass all of the objects we are interested in we are not able to evaluate our pipeline using a recognition network learned on a large dataset without some finetuning.

\section{Dataset Acquisition}
\label{sec:create_dataset}

A secondary application of our region proposal method is data acquisition.  Developing new datasets suitable for training deep learning models is normally a heavily human-time intensive process (\cite{imagenet_cvpr09}).  This is particularly important for robotics applications, where there is a dearth of pre-trained recognition models.  Using our object localization method, researchers can quickly and easily create labeled data for objects not commonly found in circulated datasets.  Our approach is described in Algorithm~\ref{algo:data_ac}.  This algorithm works by employing the use of Algorithm~\ref{algo:region} on the set of object classes of interest.  By placing an instance of a known object class in the view of the RGB-D sensor (Alg.~\ref{algo:data_ac}, Line 4), we can store the streaming output of Algorithm~\ref{algo:region} along with the user-provided label (Alg.~\ref{algo:data_ac}, Lines 5-7) in a supervised learning dataset.  This process is repeated for the full set of objects that a user is interested in at multiple locations throughout the scene.  As Algorithm~\ref{algo:region} is very fast, it is possible to store a large quantity of data very quickly.

\begin{figure}[h]
	\centering
	\fcolorbox{white}{white}{\includegraphics[width=0.09\textwidth]{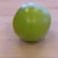}}
	\fcolorbox{white}{white}{\includegraphics[width=0.09\textwidth]{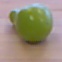}}
	\fcolorbox{white}{white}{\includegraphics[width=0.09\textwidth]{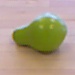}}
	\fcolorbox{white}{white}{\includegraphics[width=0.09\textwidth]{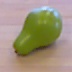}}
	
	\fcolorbox{white}{white}{\includegraphics[width=0.09\textwidth]{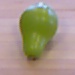}}
	\fcolorbox{white}{white}{\includegraphics[width=0.09\textwidth]{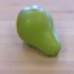}}
	\fcolorbox{white}{white}{\includegraphics[width=0.09\textwidth]{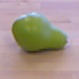}}
	\fcolorbox{white}{white}{\includegraphics[width=0.09\textwidth]{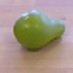}}
    \caption{Example data captured using our object localization procedure with multiple orientations.}
    \label{fig:data_ac}
\end{figure}

\begin{algorithm}[h]
\caption{Dataset Acquisition}
\begin{algorithmic}[1]
\State $\textbf{Given }\textnormal{object class labels, }\mathbb{Y}_{labels}$
\State $\textbf{Init }\mathbb{X}_{data} \gets \emptyset, \mathbb{Y}_{data} \gets \emptyset$
\For {$y_{label} \in \mathbb{Y}_{labels}$}
\State $\textnormal{place object of type } y_{label} \textnormal{ in the scene} $
\State $\mathbb{B},\mathbb{P} \gets \textnormal{Algorithm~\ref{algo:region}(object point cloud)}$
\State $b_{rgb} \gets \textnormal{Convert } \mathbb{B} \textnormal{ to RGB space}$
\State $x \gets \textnormal{image patch defined by } b_{rgb}$
\State $\mathbb{X}_{data} \gets \{\mathbb{X}_{data} \cup x\}$
\State $\mathbb{Y}_{data} \gets \{\mathbb{Y}_{data} \cup y_{label}\}$
\EndFor
\State $\textbf{return} \hspace{2mm} \mathbb{X}_{data}, \mathbb{Y}_{data}$
\end{algorithmic}
\label{algo:data_ac}
\end{algorithm}


While capturing example images, it helps the model to generalize if the position and orientation of object(s) are altered thereby providing multiple views of each class.  It can also help to alter aspects like lighting conditions and out-of-plane rotations.  An example of the types of data collected via this method can be seen in Figure~\ref{fig:data_ac}.  Source code for the dataset acquisition process is a part of the released ROS package (Extension~\ref{sec:code}).

\section{Experimental Design}
\label{sec:exp}

To evaluate the efficacy of our system, we analyze the accuracy of our approach in localizing and identifying objects in a variety of realistic tabletop scenarios.  We begin by demonstrating the dataset building capabilities of our system (Section~\ref{subsec:data}), which allows us to train our own CNN model from scratch and finetune three well known architectures whose weights were pre-trained on the ImageNet dataset (Section~\ref{subsec:models}).  We then evaluate our object detection pipeline on 40 realistic tabletop scenes.  By evaluating our pipeline with four different recognition models, we demonstrate the ease with which one can update the underlying classification model to stay in line with the state-of-the-art.  We compare the accuracy of the different models to examine the effect of different CNN architectures on the efficacy of our pipeline (Section~\ref{subsec:acc}).


\subsection{Object Dataset}
\label{subsec:data}

\begin{figure}[b]
	\centering
	\includegraphics[width=0.45\textwidth]{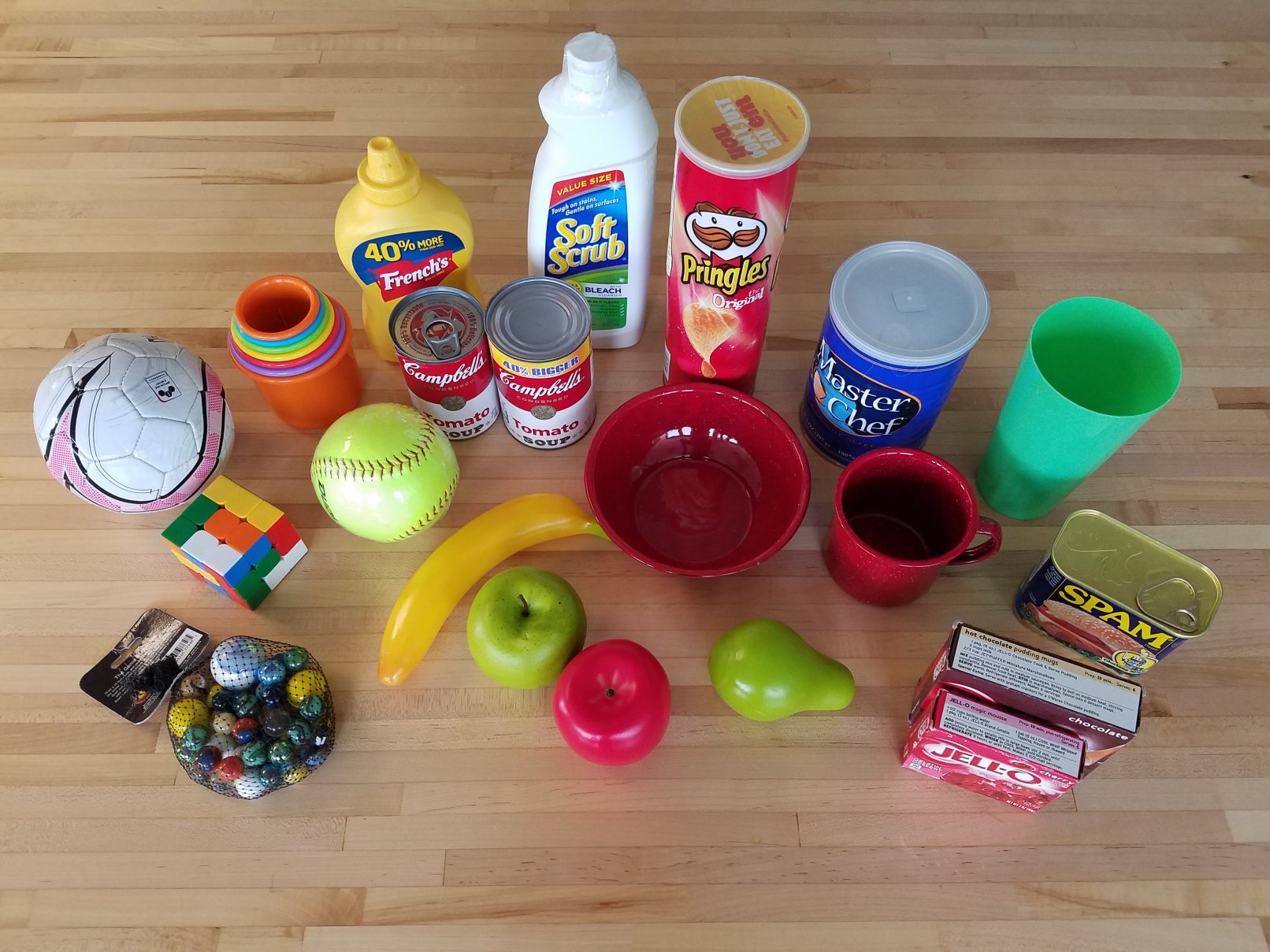}
	\caption{Object set used to test detection pipeline. 22 object instances and 19 object classes in total.  YCB Food: mustard, soup (x2), pringles, ground coffee, spam, jello (x2), apple (x2), pear, banana.  YCB Kitchen: mug, bowl, bleach.  YCB Shape: marbles, rubiks cube, soccer ball, softball, toy.  Other: cup.}
	\label{fig:data}
\end{figure}

We began by creating an object dataset consisting of 19 object classes and 22 total object instances.  The objects were almost exclusively taken from the YCB object dataset (\cite{calli2015benchmarking}).  The specific objects chosen were relevant to our target domain: namely, common household items that a user of a robotic arm may wish to interact with.  The full object set can be seen in Figure~\ref{fig:data}.  The dataset we collected consists of a total of 2640 images split evenly by class.  

\subsection{Model Architectures}
\label{subsec:models}

We trained and tested four different CNN architectures.  The first is a small model that we trained from scratch.  The other three are well tested architectures that we initialized with weights learned from the ImageNet dataset and finetuned on our own dataset.  To train each model we used an $80/20\%$ train/test split of our dataset.  During training of each network we also used different forms of data augmentation including random rotations, width and height shifts, shear mapping, and horizontal and vertical flipping.  All models are implemented using the Keras library (\cite{chollet2015}).

\subsubsection{Small Model}

The first network architecture that we evaluated is a small 6 layer convolutional neural network.  The input layer is connected to a sequence of 3 convolutional layers with 3x3 filters.  Each layer uses the ReLu non-linear activation function and is followed by 2x2 max pooling.  This sequence is then followed by 3 additional layers of non-convolutional filters.  During training, dropout is applied after the first two of the fully connected layers.  The output layer is a learned soft-max classifier.  

\subsubsection{VGG-16}

The second network architecture that we evaluated is the VGG-16 network developed by \cite{Simonyan15}.  This network has 16 layers and was the first published work to use very small 3x3 convolutional filters.  One of the key insights of this work was that sequences of small convolutional filters are capable of representing higher-order features otherwise captured by larger (more computationally expensive) receptive fields, like 7x7 or 9x9 filters.  This network has previously been demonstrated to work well on many object recognition datasets.

\subsubsection{Inception Network}

The third network architecture that we evaluated is the Inception v3 network developed by \cite{szegedy2015rethinking}.  This network has 22 layers and is another popular architecture that that places a specific focus on reducing the necessary computation at test time to improve the model's efficiency.  In this work, the authors propose the parallel computation of pooling, 1x1 and 3x3 filters which are then combined into a concatenated vector space (known as an inception module).  By using 1x1 convolutions to reduce the filter space before computing the relatively more expensive 3x3 convolutions, the authors are able to reduce the computational complexity of this operation while improving the overall performance of the model. 

\begin{figure*}[t]
	\centering
	
	\fcolorbox{white}{white}{\subfloat[]{\label{ex_04}\includegraphics[width=0.32\textwidth]{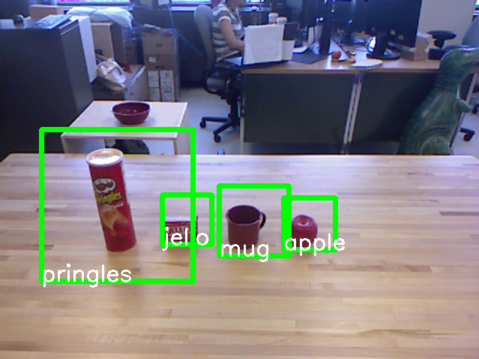}}}
	\fcolorbox{white}{white}{\subfloat[]{\label{ex_05}\includegraphics[width=0.32\textwidth]{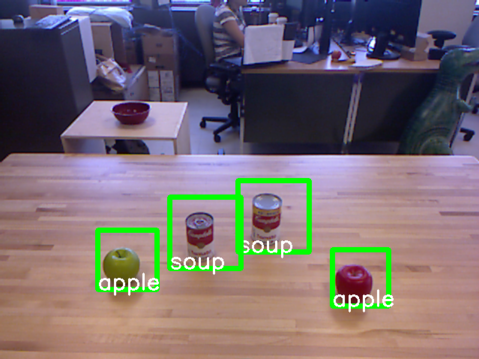}}}
	\fcolorbox{white}{white}{\subfloat[]{\label{ex_06}\includegraphics[width=0.32\textwidth]{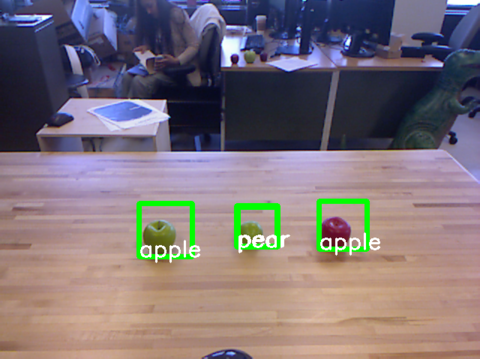}}}
	
	\fcolorbox{white}{white}{\subfloat[]{\label{ex_01}\includegraphics[width=0.32\textwidth]{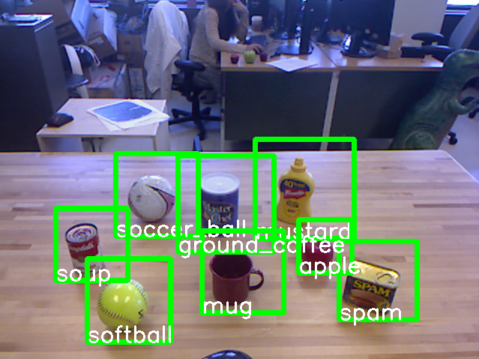}}}
	\fcolorbox{white}{white}{\subfloat[]{\label{ex_02}\includegraphics[width=0.32\textwidth]{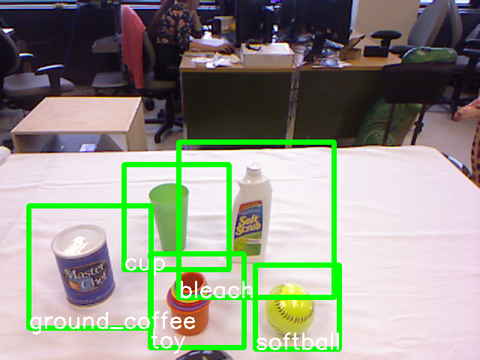}}}
	\fcolorbox{white}{white}{\subfloat[]{\label{ex_03}\includegraphics[width=0.32\textwidth]{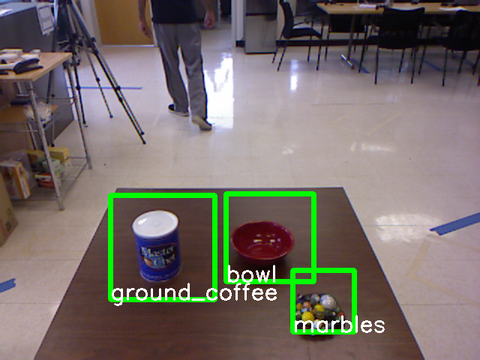}}}
	
	\caption{Six example scenes from our experimental testing set with various object configurations and environments. These scenes demonstrate the efficacy and highlight the generalizability of our approach.  (a) Four objects, all a similar color. (b) Two distinct instances of two different classes. (c) Two distinct but very similar classes --- note, the long edge of the pear is hidden.  (d) Eight objects in various configurations. (e) Five objects on a white tablecloth. (f) Three objects on a dark brown table.  The scenes in (e) and (f) are tested on tables distinct from the table used in the original data collection.}
	\label{fig:example_output}
\end{figure*}

\subsubsection{Residual Network}

The fourth, and last, network architecture that we evaluated is the Resdiual network proposed by \cite{he2015deep}.  The standard implementation of this network is 152 layers deep and the result of this work is a network structure that allows one to train much deeper networks.  In particular, the main insight of this work is the idea of \textit{skip connections}.  That is, instead of connecting the output of each layer to each following layer sequentially, one connects the output of each layer to the layer \textit{after} the next layer.  The concept behind this architecture is to encourage the network to learn \textit{residual updates} from one layer to the next.  In their paper, \cite{he2015deep} demonstrate that naively adding more and more layers to a network does not necessarily improve the performance of the network, while using residual connections dramatically improves the results.
\\
\\
All four models were successfully trained in less than 50 epochs with a batch size of 32.  By observing the validation loss during training, it was clear that the pre-trained networks learned significantly quicker than the model trained from scratch.  However, while all networks can run efficiently during test-time on a mid-tier GPU (nVidia 860m), only the smaller network can be trained on this GPU.  Finetuning the larger networks requires a more powerful computer and GPU --- during training we used an nVidia Titan X.  We then transfered the network weights for these models to the less powerful computer for the experiments.

\subsection{Evaluation Scenes}
\label{subsec:acc}

To analyze the accuracy of our pipeline, we developed 40 realistic tabletop scenes with varying numbers of objects, object configurations, clutter and backgrounds.  The number of objects in a scene ranges from three to eight.  In all evaluation scenes, the objects themselves are the same physical objects used to collect the training data, however they are collected separately and we randomize each object's position and orientation in the evaluation scenes.  Of the 40 different scenes, we include 20 in the same environment as the initial data collection, with random object configurations (Scenes 1-20). We then evaluate five in the same environment with the addition of a white tablecloth to hide the original table top surface (Scenes 21-25) and five on a new table with a much darker tabletop (Scenes 26-30).  These two sets of tabletop scenes demonstrate the generalizability of the recognition models to novel environments.  The final 10 scenes are made up of five scenes collected with a moving camera where the camera beginning on the left side of the environment and moving towards the right side (Scenes 31-35) and five with a moving camera with the camera beginning high up in the environment and slowly moving down (Scenes 36-40).  This final set of experiments is particularly relevant to mobile robotics where the platform may be moving.  The full set of 40 test scenes can be seen with their descriptions in Appendix~\ref{sec:appendix}.  Six example scenes can be seen in Figure~\ref{fig:example_output}.  Our experimental scenes are similar to those released by \cite{lai2011large}, however, we develop a larger number of scenes for testing and specifically focus on scenes that demonstrate particular capabilities of our approach (e.g. larger number of objects in a scene, invariance to object features like color, more cluttered environments and a variety of backgrounds).

\begin{table*}[t]
\begin{tabular}{l | c c c c c c c c c c c c c c c}
\multicolumn{16}{c}{\bfseries Average Recognition Accuracy} \\
Model & 1 & 2 & 3 & 4 & 5 & 6 & 7 & 8 & 9 & 10 & 11 & 12 & 13 & 14 & 15 \\
\hline
Small & 1.00 & 1.00 & \textit{0.75} & \textit{0.96} & \textit{0.95} & 1.00 & 1.00 & 1.00 & 1.00 & 1.00 & \textit{0.33} & \textit{0.83} & 1.00 & 1.00 & 1.00\\
VGG16 & 1.00 & \textit{0.75} & \textit{0.99} & 1.00 & \textit{0.67} & 1.00 & 1.00 & 1.00 & 1.00 & 1.00 & 1.00 & 1.00 & 1.00 & 1.00 & 1.00\\
Inception & 1.00 & 1.00 & 1.00 & 1.00 & \textit{0.46} & 1.00 & 1.00 & 1.00 & 1.00 & 1.00 & 1.00 & 0.91 & \textit{0.85} & 1.00 & 1.00\\
Residual & 1.00 & 1.00 & 1.00 & 1.00 & \textit{0.66} & 1.00 & 1.00 & 1.00 & 1.00 & 1.00 & 1.00 & \textit{0.95} & \textit{0.67} & 1.00 & 1.00 \\
\end{tabular}
\newline
\vspace{2mm}
\newline
\begin{tabular}{l | c c c c c c c c c c c c c c c}
Model & 16 & 17 & 18 & 19 & 20 & 21 & 22 & 23 & 24 & 25 & 26 & 27 & 28 & 29 & 30 \\
\hline
Small & \textit{0.67} & 1.00 & 1.00 & 1.00 & 1.00 & \textit{0.75} & 1.00 & \textit{0.96} & \textit{0.80} & \textit{0.75} & \textit{0.33} & \textit{0.33} & 1.00 & \textit{0.38} & \textit{0.67}\\
VGG16 & 1.00 & 1.00 & 1.00 & 1.00 & 1.00 & 1.00 & \textit{0.95} & \textit{0.74} & \textit{0.80} & 1.00 & 1.00 & 1.00 & \textit{0.79} & 1.00 & 1.00\\
Inception & 1.00 & 1.00 & \textit{0.67} & 1.00 & 1.00 & \textit{0.98} & \textit{0.96} & \textit{0.37} & \textit{0.80} & \textit{0.95} & \textit{0.83} & \textit{0.71} & 1.00 & \textit{0.95} & \textit{0.93}\\
Residual & 1.00 & 1.00 & \textit{0.75} & 1.00 & 1.00 & 1.00 & 1.00 & \textit{0.75} & \textit{0.80} & 1.00 & 1.00 & 1.00 & \textit{0.67} & 1.00 & 1.00\\
\end{tabular}
\newline
\vspace{2mm}
\newline
\begin{tabular}{l | c c c c c c c c c c | c }
Model & 31 & 32 & 33 & 34 & 35 & 36 & 37 & 38 & 39 & 40 & Overall \\
\hline
Small & 1.00 & \textit{0.99} & 1.00 & \textit{0.94} & \textit{0.87} & 1.00 & \textit{0.88} & 1.00 & \textit{0.99} & 1.00 & \textit{0.88} \\
VGG16 & \textit{0.97} & 1.00 & 1.00 & 1.00 & \textit{0.99} & \textit{0.99} & 1.00 & 1.00 & 1.00 & 1.00 & \textit{0.97} \\
Inception & \textit{0.95} & 1.00 & 1.00 & \textit{0.71} & \textit{0.99} & 1.00 & 1.00 & 1.00 & 1.00 & 1.00 & \textit{0.93} \\
Residual & 1.00 & 1.00 & 1.00 & \textit{0.99} & \textit{0.97} & 1.00 & 1.00 & 1.00 & 1.00 & 1.00 & \textit{0.95} \\
\end{tabular}
\newline
\vspace{0.5mm}
\newline
\caption{Recognition Accuracy.  The average recognition of all four models on each of the 40 scenes we tested in our experiments.  The value computed for each scene is an average of the accuracy over all objects in the first 100 frames captured at 5Hz using an RGB-D camera.  Accuracy falling short of 100$\%$ is italicized.}
\label{tab:rec_acc}
\end{table*}

\section{Results}
\label{sec:results}

To evaluate the accuracy of our pipeline on the scenes described in Section~\ref{subsec:acc}, we collect 100 continuous pointcloud frames captured at 5Hz from the RGB-D camera and compare the output of our pipeline to hand-labeled ground truth.  Our experiments were run on a Core i7 laptop with a mid-tier mobile GPU (nVidia GeForce 860M).  We compute the average accuracy of the pipeline by taking the mean of the accuracy over those 100 frames.  That is, for the pipeline to achieve 100$\%$ on a given scene, it needs to correctly predict the correct object class for \textit{each object} in \textit{each frame}.  Averaging the success of our approach over all the frames is particularly important for the scenes in which the camera is moving as there is greater variation between frames than when the camera and objects are static.  Example results can be seen in Figure~\ref{fig:example_output} and a full breakdown is presented in Table~\ref{tab:rec_acc}.

The results demonstrate that all four tested architectures are able to perform well on the 40 experimental test scenes.  Of note is the fact that these scenes were varied in the amount of clutter, in the orientation and position of the objects, in the presentation of which objects, and in some cases included novel tabletops (Appendix~\ref{sec:appendix}, Figures~\ref{scene_21}-\ref{scene_30}) and camera motion (Appendix~\ref{sec:appendix}, Figures~\ref{scene_31}-\ref{scene_40}).  In order of increasing performance, we find that the small model achieved an average accuracy of 88$\%$ on the full test set, the Inception network achieved an average accuracy of 93$\%$, the Residual network achieved an average accuracy of 95$\%$ and the VGG-16 network achieved an average accuracy of 97$\%$.  While we do observe a clear difference between the small model and the three well known models, recall that the small model is trained from scratch on a mid-tier GPU while the known models are first pre-trained on the ImageNet dataset and then further trained on a more powerful GPU.  We note that even in robotics applications that require on-board computation, it is often possible to train a model off-line on a more capable computer so long as it can later run on the on-board computer at test time.


To asses the suitability of our approach for use by robots in making online predictions, we also evaluate the running time of our system.  The full pipeline runs at an average of 12Hz, which is suitable for robotic manipulation tasks in our target domain (i.e. households).  


\section{Discussion}
\label{sec:disc}

The presented paradigm appears to be a promising direction for practical robotic perception systems.  The marriage of state-of-the-art techniques from robotics and computer vision helps produce a fast and accurate object detection framework that can be easily incorporated into any tabletop manipulation task.  It is a real-time system that does not require top-of-the-line hardware and produces competitive results.  This methodology is additionally useful for creating novel datasets which suggests that this approach could be useful for other researchers and advanced users alike.  

Our approach differs from related work in three main ways.  The first is that unlike image-only based approaches, we use the 3D geometric features of a scene to compute the region proposals that we then feed to our recognition model for classification.  This ensures that we only send a single image patch per object in the scene which is extremely efficient when compared to state-of-the-art image-based approaches.  The second way in which our work differs, is that unlike point-cloud based approaches that both localize and recognize the objects in the depth space, we localize points in the depth space, but recognize the objects in the image space using convolutional neural networks.  Deep learning based approaches have proven extremely effective in the computer vision community and our work demonstrates their applicability to robotics as well.  In particular, we observe 97$\%$ average precision over all scenes by the best performing model (Table~\ref{tab:rec_acc}).  Lastly, unlike with image-based approaches which locate objects only in 2D, our method produces the precise 3D position of the object with minimal additional computation.

In the remainder of the discussion we will compare our approach to image-based methods (Section~\ref{sec:sub_comp}), examine the generalizability of our system (Section~\ref{sec:sub_gen}), note specific cases of failure (Section~\ref{sec:sub_fail}) and discuss future directions (Section~\ref{sec:sub_future}).

\subsection{Comparison to Image-Based Methods}
\label{sec:sub_comp}

Ideally, our analysis would have included a direct comparison between our geometric region proposal method and state-of-the-art image-based approaches from the computer vision community
(e.g. R-CNN (\cite{girshick2014rich}), Fast R-CNN (\cite{girshick2015fast}) and Faster R-CNN (\cite{ren2015faster})). However, there are a number of key differences between our approach and this body of
work that (1) make such a direct comparison challenging and (2) highlight some of the gains of our approach (for use within robotics in particular). These differences include the required training data, available 3D information and execution speed, which are discussed next.

\subsubsection{Training Data Requirements}

One notable distinction is a significant difference in the \textit{type} and \textit{amount} of required training data.  

In our object detection pipeline, objects are localized autonomously by exploiting known geometric properties of the scene.  All that the human provides is a class label.  By comparison, Fast R-CNN and Faster R-CNN require training data that includes images labeled not only with the object class, but also with the corresponding bounding box for each object in the scene.  Each bounding box is drawn by hand---a significant human-time intensive process.  For this reason, it is unlikely that this type of data will become widely available for all objects of interest in our target domain (the home) in the near future, which prohibits the training or finetuning any of these models on novel datasets (that do not include bounding box labels).

In addition to the required localization data, the image-only based approaches rely much more heavily on the training data including a \textit{background} class, which is used to recognize false positives nominated by the region proposal method.  Unlike Fast R-CNN and Faster R-CNN, the original R-CNN approach (a significantly slower approach) does not require training a network that localizes objects in a scene.  However, likewise, it still utilizes a region proposal method that proposes (on average) significantly more object locations than there actually are objects (R-CNN and Fast R-CNN generate $\sim$2000 object proposals and Faster R-CNN generates $\leq$ 300).  To solve this problem, each of the aforementioned approaches utilize a background class which can reject the false positives.  Again, this requires collecting more training data.  Even worse, it increases the likelihood of a false positive at the end of the pipeline, whereas we saw \textit{zero} false positives (related to the existence of an object) in our experiments.


\subsubsection{3D Localization}

Another notable distinction is that approaches from the computer vision community only solve the localization problem in 2D.  This is insufficient for robotics applications as the robot itself exists in 3D and must interact with other objects in the same space.  Additionally, not only do these computer vision models lack depth information, but that information is not directly available in the datasets used to train these models (such as ImageNet), which increases the difficulty of incorporating depth into these models.  

\subsubsection{Speed at Test Time}

We furthermore compare our runtime to that of competitive approaches in speed and accuracy, the image-based R-CNN, Fast R-CNN and Faster R-CNN methods.  Comparisons were run on the same computer, with the same size images (640$\times$480), and using test set images from the same dataset used to create a model's training set. Under these conditions,  R-CNN was able to run at an average of 0.4Hz (using Selective Search  (\cite{uijlings2013selective}) for region proposals), Fast R-CNN was able to run at an average of 1.33Hz and Faster R-CNN was able to run at an average of 5Hz.  Our approach ran at an average of 12Hz, and thus demonstrates a speed up factor of 30x, 9.2x and 2.4x, respectively.

\subsection{Generalizability}
\label{sec:sub_gen}

To asses generalizability, we tested our pipeline on 40 realistic tabletop scenes covering a large variety of possible object classes and configurations.  The scenes vary in which objects are present, where they are located, their orientation and the amount of clutter (see Appendix~\ref{sec:appendix}).  We also examine the effect of a moving camera and testing on tabletops that are unique from the one used during data collection.  

\begin{figure}[b]
	\centering
	\fcolorbox{white}{white}{\includegraphics[width=0.23\textwidth]{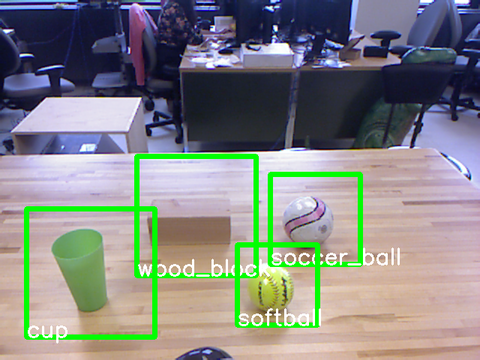}}
	\fcolorbox{white}{white}{\includegraphics[width=0.23\textwidth]{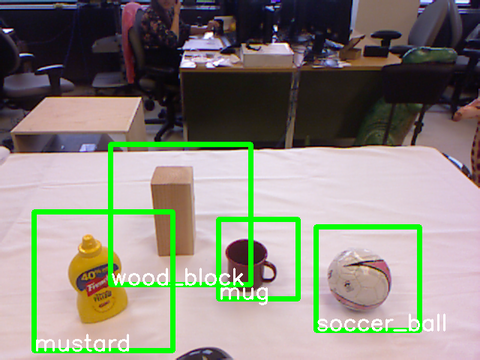}}
	\caption{In both scenes the pipeline is able to correctly localize and recognize all objects.  In particular, by observing the wood block in both scenes we see that the recognition network is capable of accounting for out-of-plane orientation changes.}
	\label{fig:diff_orientation}
\end{figure}

We highlight a few noteworthy examples.  The scene shown in Figure~\ref{ex_04} demonstrates that the recognition system is capable of distinguishing between objects of \textit{similar color}.  The scene shown in Figure~\ref{ex_05} demonstrates that the system is able to generalize to different instances of physical objects of the \textit{same class}, and the scene shown in Figure~\ref{ex_06} demonstrates that the system is able to differentiate between classes that are \textit{extremely similar}.  In particular, in this last image, notice that the pear is intentionally oriented away from the camera hiding the top half of the pear, likely the largest visually differentiating factor between a pear and an apple.  In Figure~\ref{fig:diff_orientation} we see that the recognition system is able to account for \textit{out-of-plane orientation} changes (note the orientation of the wood block in both images).

We also demonstrate that the system performs well when there are \textit{many objects} in the scene (Figure~\ref{ex_01}) and when the tabletop is distinct in appearance from the tabletop used to capture the training data (Figures~\ref{ex_02} and ~\ref{ex_03} as well as Appendix~\ref{sec:appendix}, Figures~\ref{scene_21}-\ref{scene_30}).  The system moreover performs well when the \textit{camera is moving} during the data capture, demonstrating that this approach is viable for mobile robots (Appendix~\ref{sec:appendix}, Figures~\ref{scene_31}-\ref{scene_40}).  In the first five of these scenes, the camera was held at a constant height and was moved about one meter from the left side of the scene to the right.  In the second five of these scenes, the camera was moved about one meter from the top of the scene to the bottom.

\subsection{Failures}
\label{sec:sub_fail}

\begin{figure}[h]
	\centering
	\fcolorbox{white}{white}{\includegraphics[width=0.1\textwidth]{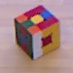}}
	\fcolorbox{white}{white}{\includegraphics[width=0.1\textwidth]{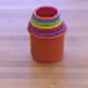}}
	\caption{Visually similar objects.  Left: Rubiks cube. Right: Toy.}
	\label{fig:rc_toy}
\end{figure}

In the cases of recognition failure, it was often fairly clear why there was a misclassification.  For example, one of the more common mis-labelings was the toy and rubiks cube which have similar color schemes (Figure~\ref{fig:rc_toy}).  However, it is clear from the results that these small errors can be easily overcome by improved modeling techniques.  For example, Scenes 16, 25 and 35 each include either the toy or rubiks cube and our results show that the small model is unable to correctly identify these objects.  However, the three models pre-trained on the (larger) ImageNet dataset are able to correctly classify all objects in each of these scenes (Table~\ref{tab:rec_acc}).

\noindent Additionally, when reviewing the accuracy of these models over the entire test set, we see a large improvement when we move from the small model trained from scratch to the pre-trained models that we finetuned.  In particular it appears clear that finetuning a pre-trained network helps with generalization to different environments.  For example, we see a significantly larger effect of the background color on the small model (where the average accuracy of the model drops to 70$\%$) than any of the pre-trained networks (e.g. the accuracy of the VGG16 only drops to 93$\%$).  However, the trade-off is that the three larger pre-trained networks needed to be finetuned on the more powerful and more expensive Titan X GPU. 




One point of failure is how the recognition model handles objects that were not in the training set.  Our current approach will choose the most likely class label as defined by the probabilities that we get from the softmax output layer of each network.  However, this choice is not well suited under the open set world assumption where we expect to see novel objects that were not included in the training data.  Instead a robot needs to be aware of when it comes across a novel object.  To solve this problem, we can incorporate statistical techniques for detecting class outliers and incorporating novel objects as described in~\cite{bendale2015towards}.


\subsection{Future Directions}
\label{sec:sub_future}

We expect that highly cluttered environments will require improved segmentation approaches in the point cloud representation.  In particular, it is likely that depth-only segmentation will fail in scenes where objects actually sit on top of one another, for example, imagine objects sitting on a bookshelf.  A potential area of further work in this research would be to combine the current depth-based segmentation with image-based segmentation approaches to incorporate color information as well into the segmentation procedure.

An additional area of possible improvement and refinement in our system would be to include the depth information from the RGB-D camera as a fourth channel in our CNN architecture.  Similar to the benefit of using color information in the segmentation process, the recognition portion of our pipeline could be improved by incorporating depth information into the model.  However, at least for now, this would reduce our ability to finetune pre-trained networks (a cheap and efficient way to use features learned from larger datasets) as currently these networks only include color information.

\section{Conclusion} 
\label{sec:conclusion}

In this paper, we describe and demonstrate a simple, and fast, object detection pipeline for tabletop manipulation tasks using robot vision.  We validate the efficacy of our approach with a thorough study to test the speed, accuracy and generalizability of our method.  We found our system to be capable of running in real-time (12Hz) on limited-capability hardware.  The described system owes its speed and computational efficiency to the minimal set of regions proposed through unsupervised methods of analysis in the point cloud space.  We also demonstrate that our method makes it easy to collect novel datasets which can be used to train recognition models from scratch or used to finetune models pre-trained on larger datasets.  The modular design of the pipeline makes it easy to incorporate new recognition models in order to stay in-line with the state-of-the-art.  In our experiments, we found that incorporating the state-of-the-art CNN architectures allowed us to achieve a 97$\%$ detection accuracy on our varied experimental dataset.  Our approach owes it accuracy and generalizability in the recognition space to Convolutional Neural Networks.  


In future work we hope to improve the capabilities of our approach by incorporating color information into our segmentation approach and incorporating depth information into our recognition models.  We also plan to demonstrate how this approach can be used under the open-set assumption.  We believe that integrating these changes will also allow us to expand the classes of scenes in which this approach can be validated.

The code for both the full object detection pipeline as well as the dataset acquisition portion can be found at \url{https://github.com/asbroad/geom_rcnn} and are included with an open-source MIT license in Extension~\ref{sec:code}.

\section*{Acknowledgments}
This material is based upon work supported by the National Science Foundation under Grant CNS 1329891. Any opinions, findings and conclusions or recommendations expressed in this
material are those of the authors and do not necessarily reflect the views of the aforementioned institutions.

\bibliographystyle{plainnat}
\bibliography{refs}

\begin{thebibliography}{39}
\providecommand{\natexlab}[1]{#1}
\providecommand{\url}[1]{\texttt{#1}}
\expandafter\ifx\csname urlstyle\endcsname\relax
  \providecommand{\doi}[1]{doi: #1}\else
  \providecommand{\doi}{doi: \begingroup \urlstyle{rm}\Url}\fi

\bibitem[Bendale and Boult(2015)]{bendale2015towards}
Abhijit Bendale and Terrance Boult.
\newblock
  \href{http://www.cv-foundation.org/openaccess/content_cvpr_2015/papers/Bendale_Towards_Open_World_2015_CVPR_paper.pdf}{Towards
  Open World Recognition}.
\newblock In \emph{Proceedings of the IEEE Conference on Computer Vision and
  Pattern Recognition (CVPR)}, pages 1893--1902, 2015.

\bibitem[Calli et~al.(2015)Calli, {Walsman}, Singh, Srinivasa, Abbeel, and
  Dollar]{calli2015benchmarking}
Berk Calli, Aaron {Walsman}, Arjun Singh, Siddhartha Srinivasa, Pieter Abbeel,
  and Aaron Dollar.
\newblock \href{http://ieeexplore.ieee.org/document/7254318/}{Benchmarking in
  Manipulation Research: The {YCB} Object and Model Set and Benchmarking
  Protocols}.
\newblock In \emph{IEEE Robotics and Automation Magazine}, August 2015.

\bibitem[Chen and Yuille(2005)]{chen2005time}
Xiangrong Chen and Alan~L Yuille.
\newblock \href{http://dl.acm.org/citation.cfm?id=1099539.1099929}{A
  Time-Efficient Cascade for Real-Time Object Detection: With Applications for
  the Visually Impaired}.
\newblock In \emph{Proceedings of the IEEE Conference on Computer Vision and
  Pattern Recognition (CVPR) Workshops}, pages 28--28, 2005.

\bibitem[Chen et~al.(2015)Chen, Kundu, Zhu, Berneshawi, Ma, Fidler, and
  Urtasun]{chen20153d}
Xiaozhi Chen, Kaustav Kundu, Yukun Zhu, Andrew~G Berneshawi, Huimin Ma, Sanja
  Fidler, and Raquel Urtasun.
\newblock
  \href{https://papers.nips.cc/paper/5644-3d-object-proposals-for-accurate-object-class-detection}
  {{3D} Object Proposals for Accurate Object Class Detection}.
\newblock In \emph{Proceedings of Advances in Neural Information Processing
  Systems (NIPS)}, pages 424--432, 2015.

\bibitem[Chollet(2015)]{chollet2015}
François Chollet.
\newblock keras.
\newblock \url{https://github.com/fchollet/keras}, 2015.

\bibitem[Couprie et~al.(2014)Couprie, Farabet, Najman, and
  LeCun]{couprie2013indoor}
Camille Couprie, Clement Farabet, Laurent Najman, and Yann LeCun.
\newblock \href{http://jmlr.org/papers/v15/couprie14a.html}{Convolutional Nets
  and Watershed Cuts for Real-Time Semantic Labeling of RGBD Videos}.
\newblock \emph{Journal of Machine Learning Research}, 15:\penalty0 3489--3511,
  2014.

\bibitem[Dahan et~al.(2012)Dahan, Chen, Shamir, and
  Cohen-Or]{dahan2012combining}
Meir~Johnathan Dahan, Nir Chen, Ariel Shamir, and Daniel Cohen-Or.
\newblock
  \href{http://link.springer.com/article/10.1007/s00371-011-0667-7}{Combining
  Color and Depth for Enhanced Image Segmentation and Retargeting}.
\newblock \emph{The Visual Computer}, 28\penalty0 (12):\penalty0 1181--1193,
  2012.

\bibitem[Deng et~al.(2009)Deng, Dong, Socher, Li, Li, and
  Fei-Fei]{imagenet_cvpr09}
Jia Deng, Wei Dong, Richard Socher, Li-Jia Li, Kai Li, and Li~Fei-Fei.
\newblock
  \href{http://ieeexplore.ieee.org/document/5206848/?arnumber=5206848&tag=1}{{ImageNet}:
  A large-scale hierarchical image database}.
\newblock In \emph{Proceedings of the IEEE Conference on Computer Vision and
  Pattern Recognition (CVPR)}, pages 248--255, 2009.

\bibitem[Girshick(2015)]{girshick2015fast}
Ross Girshick.
\newblock
  \href{http://www.cv-foundation.org/openaccess/content_iccv_2015/papers/Girshick_Fast_R-CNN_ICCV_2015_paper.pdf}{Fast
  {R-CNN}}.
\newblock In \emph{Proceedings of the IEEE International Conference on Computer
  Vision and Pattern Recognition (CVPR)}, pages 1440--1448, 2015.

\bibitem[Girshick et~al.(2014)Girshick, Donahue, Darrell, and
  Malik]{girshick2014rich}
Ross Girshick, Jeff Donahue, Trevor Darrell, and Jitendra Malik.
\newblock
  \href{http://www.cv-foundation.org/openaccess/content_cvpr_2014/html/Girshick_Rich_Feature_Hierarchies_2014_CVPR_paper.html}{Rich
  Feature Hierarchies for Accurate Object Detection and Semantic Segmentation}.
\newblock In \emph{Proceedings of the IEEE Conference on Computer Vision and
  Pattern Recognition (CVPR)}, pages 580--587, 2014.

\bibitem[Gupta et~al.(2014)Gupta, Girshick, Arbel\'{a}ez, and
  Malik]{gupta14rcnndepth}
Saurabh Gupta, Ross Girshick, Pablo Arbel\'{a}ez, and Jitendra Malik.
\newblock
  \href{http://link.springer.com/chapter/10.1007%2F978-3-319-10584-0_23}{Learning
  Rich Features from {RGB-D} Images for Object Detection and Segmentation}.
\newblock In \emph{Proceedings of the European Conference on Computer Vision
  ({ECCV})}, pages 345--360, 2014.

\bibitem[He et~al.(2015)He, Zhang, Ren, and Sun]{he2015deep}
Kaiming He, Xiangyu Zhang, Shaoqing Ren, and Jian Sun.
\newblock
  \href{http://www.cv-foundation.org/openaccess/content_cvpr_2016/papers/He_Deep_Residual_Learning_CVPR_2016_paper.pdf}{Deep
  residual learning for image recognition}.
\newblock In \emph{Proceedings of the IEEE Conference on Computer Vision and
  Pattern Recognition (CVPR)}, 2015.

\bibitem[Hosang et~al.(2014)Hosang, Benenson, and Schiele]{hosang2014good}
Jan Hosang, Rodrigo Benenson, and Bernt Schiele.
\newblock \href{http://www.bmva.org/bmvc/2014/files/paper082.pdf}{How Good are
  Detection Proposals, Really?}
\newblock In \emph{Proceedings of the British Machine Vision Conference
  (BMVC)}, 2014.

\bibitem[Jiang and Learned-Miller(2016)]{jiang2016face}
Huaizu Jiang and Erik Learned-Miller.
\newblock \href{https://arxiv.org/pdf/1606.03473v1.pdf}{Face detection with the
  faster R-CNN}.
\newblock \emph{arXiv:1606.03473}, 2016.

\bibitem[Karan(2015)]{karan2015calibration}
Branko Karan.
\newblock
  \href{http://www.mas.bg.ac.rs/_media/istrazivanje/fme/vol43/1/8_bkaran.pdf}{Calibration
  of Kinect-type {RGB-D} sensors for robotic applications}.
\newblock \emph{FME Transactions}, 43\penalty0 (1):\penalty0 47--54, 2015.

\bibitem[Karpathy and Fei-Fei(2015)]{karpathy2015deep}
Andrej Karpathy and Li~Fei-Fei.
\newblock \href{http://cs.stanford.edu/people/karpathy/cvpr2015.pdf}{Deep
  Visual-Semantic Alignments for Generating Image Descriptions}.
\newblock In \emph{Proceedings of the IEEE Conference on Computer Vision and
  Pattern Recognition (CVPR)}, pages 3128--3137, 2015.

\bibitem[Khoshelham and Elberink(2012)]{khoshelham2012accuracy}
Kourosh Khoshelham and Sander~Oude Elberink.
\newblock \href{http://www.mdpi.com/1424-8220/12/2/1437}{Accuracy and
  Resolution of Kinect Depth Data for Indoor Mapping Applications}.
\newblock \emph{Sensors}, 12\penalty0 (2):\penalty0 1437--1454, 2012.

\bibitem[Krizhevsky et~al.(2012)Krizhevsky, Sutskever, and
  Hinton]{krizhevsky2012imagenet}
Alex Krizhevsky, Ilya Sutskever, and Geoffrey~E Hinton.
\newblock
  \href{https://papers.nips.cc/paper/4824-imagenet-classification-with-deep-convolutional-neural-networks.pdf}{Imagenet
  Classification with Deep Convolutional Neural Networks}.
\newblock In \emph{Proceedings of Advances in Neural Information Processing
  Systems}, pages 1097--1105, 2012.

\bibitem[Lai et~al.(2011)Lai, Bo, Ren, and Fox]{lai2011large}
Kevin Lai, Liefeng Bo, Xiaofeng Ren, and Dieter Fox.
\newblock
  \href{https://homes.cs.washington.edu/~xren/publication/lai_icra11_rgbd_dataset.pdf}{A
  Large-Scale Hierarchical Multi-View {RGB-D} Object Dataset}.
\newblock In \emph{Proceedings of the IEEE International Conference on Robotics
  and Automation (ICRA)}, pages 1817--1824, 2011.

\bibitem[LeCun et~al.(2015)LeCun, Bengio, and Hinton]{lecun2015deep}
Yann LeCun, Yoshua Bengio, and Geoffrey Hinton.
\newblock
  \href{http://www.nature.com/nature/journal/v521/n7553/full/nature14539.html}{Deep
  learning}.
\newblock \emph{Nature}, 521\penalty0 (7553):\penalty0 436--444, 2015.

\bibitem[Mur-Artal et~al.(2015)Mur-Artal, Montiel, and Tardos]{mur2015orb}
Raul Mur-Artal, JMM Montiel, and Juan~D Tardos.
\newblock \href{http://ieeexplore.ieee.org/document/7219438/}{{ORB-SLAM}: A
  Versatile and Accurate Monocular {SLAM} System}.
\newblock \emph{IEEE Transactions on Robotics}, 31\penalty0 (5):\penalty0
  1147--1163, 2015.

\bibitem[Pillai and Leonard(2015)]{pillai_rss15}
Sudeep Pillai and John Leonard.
\newblock \href{http://www.roboticsproceedings.org/rss11/p34.pdf}{Monocular
  {SLAM} Supported Object Recognition}.
\newblock In \emph{Proceedings of Robotics: Science and Systems (RSS)}, July
  2015.

\bibitem[Ren et~al.(2015)Ren, He, Girshick, and Sun]{ren2015faster}
Shaoqing Ren, Kaiming He, Ross Girshick, and Jian Sun.
\newblock
  \href{https://papers.nips.cc/paper/5638-faster-r-cnn-towards-real-time-object-detection-with-region-proposal-networks.pdf}{Faster
  {R-CNN}: Towards Real-Time Object Detection with Region Proposal Networks}.
\newblock In \emph{Proceedings of Advances in Neural Information Processing
  Systems (NIPS)}, pages 91--99, 2015.

\bibitem[Rusu et~al.(2009{\natexlab{a}})Rusu, Blodow, and Beetz]{rusu2009fast}
Radu~Bogdan Rusu, Nico Blodow, and Michael Beetz.
\newblock
  \href{http://ieeexplore.ieee.org/stamp/stamp.jsp?arnumber=5152473}{Fast Point
  Feature Histograms ({FPFH}) for {3D} Registration}.
\newblock In \emph{Proceedings of the IEEE International Conference on Robotics
  and Automation (ICRA)}, pages 3212--3217, 2009{\natexlab{a}}.

\bibitem[Rusu et~al.(2009{\natexlab{b}})Rusu, Blodow, Marton, and
  Beetz]{rusu2009close}
Radu~Bogdan Rusu, Nico Blodow, Zoltan~Csaba Marton, and Michael Beetz.
\newblock
  \href{http://ieeexplore.ieee.org/stamp/stamp.jsp?arnumber=5354683}{Close-Range
  Scene Segmentation and Reconstruction of {3D} Point Cloud Maps for Mobile
  Manipulation in Domestic Environments}.
\newblock In \emph{Proceedings of the IEEE/RSJ International Conference on
  Intelligent Robots and Systems (IROS)}, pages 1--6, 2009{\natexlab{b}}.

\bibitem[Rusu et~al.(2010)Rusu, Bradski, Thibaux, and Hsu]{rusu2010fast}
Radu~Bogdan Rusu, Gary Bradski, Romain Thibaux, and John Hsu.
\newblock
  \href{http://ieeexplore.ieee.org/stamp/stamp.jsp?arnumber=5651280}{Fast {3D}
  Recognition and Pose using the Viewpoint Feature Histogram}.
\newblock In \emph{Proceedings of the IEEE/RSJ International Conference on
  Intelligent Robots and Systems (IROS)}, pages 2155--2162, 2010.

\bibitem[Saxena et~al.(2009)Saxena, Sun, and Ng]{saxena2009make}
Ashutosh Saxena, Min Sun, and Andrew~Y. Ng.
\newblock
  \href{http://ieeexplore.ieee.org/stamp/stamp.jsp?arnumber=4531745}{{Make3D}:
  Learning {3D} Scene Structure from a Single Still Image}.
\newblock \emph{IEEE Transactions on Pattern Analysis and Machine Intelligence
  (PAMI)}, 31\penalty0 (5):\penalty0 824--840, 2009.

\bibitem[Schroff et~al.(2015)Schroff, Kalenichenko, and
  Philbin]{schroff2015facenet}
Florian Schroff, Dmitry Kalenichenko, and James Philbin.
\newblock
  \href{http://ieeexplore.ieee.org/abstract/document/7298682/?section=abstract}{Facenet:
  A Unified Embedding for Face Recognition and Clustering}.
\newblock In \emph{Proceedings of the IEEE Conference on Computer Vision and
  Pattern Recognition}, pages 815--823, 2015.

\bibitem[Simonyan and Zisserman(2015)]{Simonyan15}
Karen Simonyan and Andrew. Zisserman.
\newblock \href{https://arxiv.org/abs/1409.1556}{Very Deep Convolutional
  Networks for Large-Scale Image Recognition}.
\newblock In \emph{Proceedings of International Conference on Learning
  Representations (ICLR)}, 2015.

\bibitem[Song and Xiao(2014)]{song2014sliding}
Shuran Song and Jianxiong Xiao.
\newblock
  \href{http://link.springer.com/chapter/10.1007%2F978-3-319-10599-4_41}{Sliding
  Shapes for {3D} Object Detection in Depth Images}.
\newblock In \emph{Proceedings of European Conference on Computer Vision
  ({ECCV})}, pages 634--651. 2014.

\bibitem[St{\"u}ckler et~al.(2013)St{\"u}ckler, Steffens, Holz, and
  Behnke]{stuckler2013efficient}
J{\"o}rg St{\"u}ckler, Ricarda Steffens, Dirk Holz, and Sven Behnke.
\newblock \href{http://dl.acm.org/citation.cfm?id=2527951}{Efficient {3D}
  Object Perception and Grasp Planning for Mobile Manipulation in Domestic
  Environments}.
\newblock \emph{Robotics and Autonomous Systems}, 61\penalty0 (10):\penalty0
  1106--1115, 2013.

\bibitem[Szegedy et~al.(2013)Szegedy, Toshev, and Erhan]{szegedy2013deep}
Christian Szegedy, Alexander Toshev, and Dumitru Erhan.
\newblock
  \href{https://papers.nips.cc/paper/5207-deep-neural-networks-for-object-detection.pdf}{Deep
  Neural Networks for Object Detection}.
\newblock In \emph{Proceedings of Advances in Neural Information Processing
  Systems (NIPS)}, pages 2553--2561, 2013.

\bibitem[Szegedy et~al.(2015)Szegedy, Vanhoucke, Ioffe, Shlens, and
  Wojna]{szegedy2015rethinking}
Christian Szegedy, Vincent Vanhoucke, Sergey Ioffe, Jonathon Shlens, and
  Zbigniew Wojna.
\newblock \href{https://arxiv.org/pdf/1512.00567.pdf}{Rethinking the inception
  architecture for computer vision}.
\newblock \emph{arXiv:1512.00567}, 2015.

\bibitem[Tang et~al.(2012)Tang, Miller, Singh, and Abbeel]{tang2012textured}
Jie Tang, Stephen Miller, Arjun Singh, and Pieter Abbeel.
\newblock
  \href{https://people.eecs.berkeley.edu/~pabbeel/papers/TangMillerSinghAbbeel_ICRA2012.pdf}{A
  Textured Object Recognition Pipeline for Color and Depth Image Data}.
\newblock In \emph{Proceedings of the IEEE International Conference on Robotics
  and Automation (ICRA)}, pages 3467--3474, 2012.

\bibitem[Tombari et~al.(2010)Tombari, Salti, and Di~Stefano]{tombari2010unique}
Federico Tombari, Samuele Salti, and Luigi Di~Stefano.
\newblock
  \href{http://citeseerx.ist.psu.edu/viewdoc/download?doi=10.1.1.667.2598&rep=rep1&type=pdf}{Unique
  Signatures of Histograms for Local Surface Description}.
\newblock In \emph{Proceedings of European Conference on Computer Vision
  ({ECCV})}, pages 356--369. 2010.

\bibitem[Uijlings et~al.(2013)Uijlings, van~de Sande, Gevers, and
  Smeulders]{uijlings2013selective}
Jasper~RR Uijlings, Koen~EA van~de Sande, Theo Gevers, and Arnold~WM Smeulders.
\newblock
  \href{http://link.springer.com/article/10.1007/s11263-013-0620-5}{Selective
  Search for Object Recognition}.
\newblock \emph{International Journal of Computer Vision}, 104\penalty0
  (2):\penalty0 154--171, 2013.

\bibitem[Zhang et~al.(2016)Zhang, Lin, Liang, and He]{zhang2016faster}
Liliang Zhang, Liang Lin, Xiaodan Liang, and Kaiming He.
\newblock
  \href{http://link.springer.com/chapter/10.1007/978-3-319-46475-6_28}{Is
  Faster R-CNN Doing Well for Pedestrian Detection?}
\newblock In \emph{Proceedings of European Conference on Computer Vision
  (ECCV)}, pages 443--457, 2016.

\bibitem[Zhang(2012)]{zhang2012microsoft}
Zhengyou Zhang.
\newblock
  \href{https://www.microsoft.com/en-us/research/wp-content/uploads/2016/02/Microsoft20Kinect20Sensor20and20Its20Effect20-20IEEE20MM202012.pdf}{Microsoft
  Kinect Sensor and its Effect}.
\newblock \emph{IEEE MultiMedia}, 19\penalty0 (2):\penalty0 4--10, 2012.

\bibitem[Zitnick and Doll{\'a}r(2014)]{zitnick2014edge}
C~Lawrence Zitnick and Piotr Doll{\'a}r.
\newblock
  \href{https://www.microsoft.com/en-us/research/wp-content/uploads/2014/09/ZitnickDollarECCV14edgeBoxes.pdf}{Edge
  Boxes: Locating Object Proposals from Edges}.
\newblock In \emph{Proceedings of European Conference on Computer Vision
  ({ECCV})}. 2014.

\end{thebibliography}

\appendix

\section{Index to Multimedia Extensions}
\label{sec:code}

The multimedia extensions to this article are at: \url{www.ijrr.org}.

\begin{table}[h]
    \begin{tabular}{ c c c }
    \hline
    \footnotesize Extension & \footnotesize Media Type & \footnotesize Description \\ \hline
    \footnotesize 1 & \footnotesize Code & \multicolumn{1}{m{4.5cm}}{\footnotesize An efficient framework for 3D object detection in Robotics applications. It is specifically designed for detecting objects in table-top scenes (or similar environments with objects sitting on a dominant plane). The output of the system is a class label and 3D position for each object in the scene.}  \\
    \hline
    \end{tabular}
\end{table}

\section{Evaluation Data}
\label{sec:appendix}

\begin{figure*}
  \centering
\fcolorbox{white}{white}{\subfloat[Scene 1. Four objects: ground coffee, mug, soup, spam.]{\label{scene_01}\includegraphics[width=0.24\textwidth]{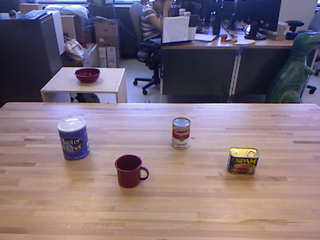}}}
\fcolorbox{white}{white}{\subfloat[Scene 2. Four objects: apple, ground coffee, mustard, toy.]{\label{scene_02}\includegraphics[width=0.24\textwidth]{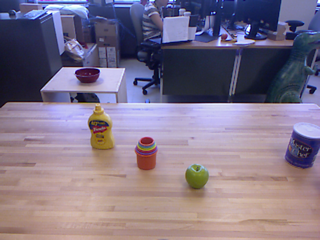}}}
\fcolorbox{white}{white}{\subfloat[Scene 3. Four objects: bleach, softball, marbles, pringles.]{\label{scene_03}\includegraphics[width=0.24\textwidth]{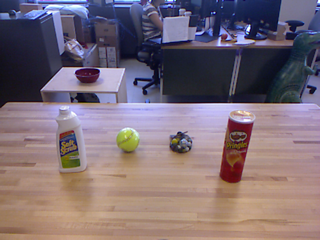}}}
\fcolorbox{white}{white}{\subfloat[Scene 4. Four objects: two different apples and soups.]{\label{scene_04}\includegraphics[width=0.24\textwidth]{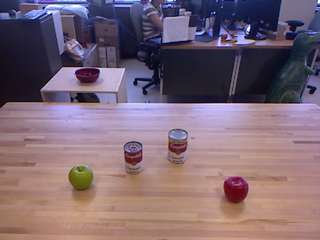}}}

\fcolorbox{white}{white}{\subfloat[Scene
5. Three objects: ground coffee, mustard, soup. Two objects are occluded.]{\label{scene_05}\includegraphics[width=0.24\textwidth]{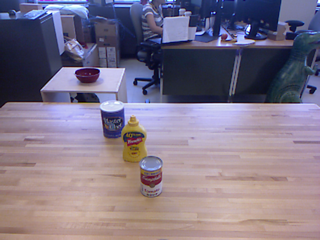}}}
\fcolorbox{white}{white}{\subfloat[Scene 6. Four objects: pringles, jellow, mug, apple.  All of which have similar colors.]{\label{scene_06}\includegraphics[width=0.24\textwidth]{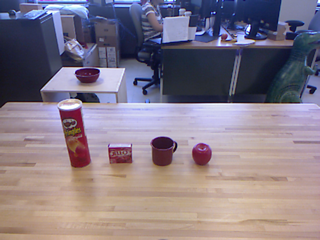}}}
\fcolorbox{white}{white}{\subfloat[Scene 7.  Three objects: two apples and a pear.  The long edge of the pear is hidden.]{\label{scene_07}\includegraphics[width=0.24\textwidth]{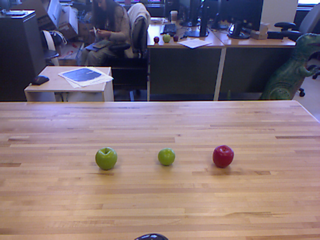}}}
\fcolorbox{white}{white}{\subfloat[Scene 8.  Three objects: soccer ball, bowl, banana.]{\label{scene_08}\includegraphics[width=0.24\textwidth]{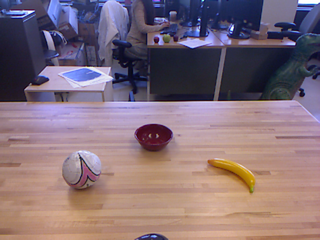}}}

\fcolorbox{white}{white}{\subfloat[Scene
9. Eight objects: soup, soccer ball, ground coffee, mustard, apple, spam, mug, softball.]{\label{scene_09}\includegraphics[width=0.24\textwidth]{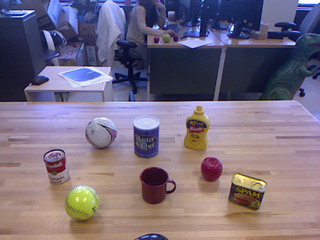}}}
\fcolorbox{white}{white}{\subfloat[Scene 10.  Three objects: wood block, cup, pear.]{\label{scene_10}\includegraphics[width=0.24\textwidth]{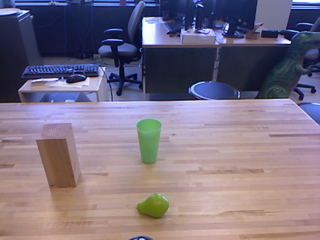}}}
\fcolorbox{white}{white}{\subfloat[Scene 11. Three objects: jello, bleach,  rubiks cube.]{\label{scene_11}\includegraphics[width=0.24\textwidth]{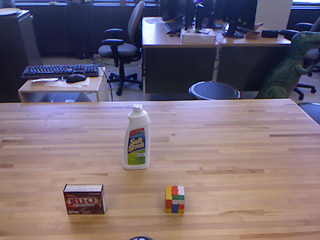}}}
\fcolorbox{white}{white}{\subfloat[Scene 12. Five objects: bowl, cup, jello, banana, marbles.]{\label{scene_12}\includegraphics[width=0.24\textwidth]{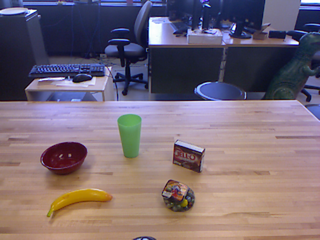}}}

\fcolorbox{white}{white}{\subfloat[Scene
13. Three objects: banana, wood block, jello.]{\label{scene_13}\includegraphics[width=0.24\textwidth]{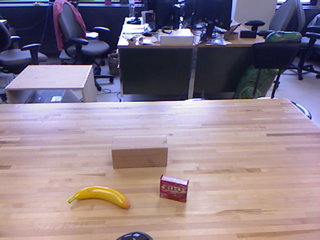}}}
\fcolorbox{white}{white}{\subfloat[Scene 14. Four objects: bowl, jello, cup, pear.]{\label{scene_14}\includegraphics[width=0.24\textwidth]{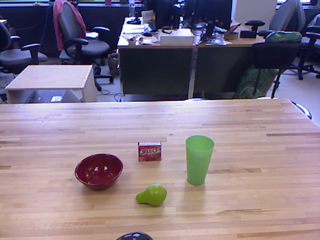}}}
\fcolorbox{white}{white}{\subfloat[Scene 15. Four objects: cup, wood block, softball, soccer ball.]{\label{scene_15}\includegraphics[width=0.24\textwidth]{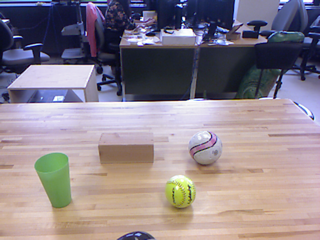}}}
\fcolorbox{white}{white}{\subfloat[Scene 16.  Three objects: rubiks cube, toy, marbles.  All objects have a similar combination of colors.]{\label{scene_16}\includegraphics[width=0.24\textwidth]{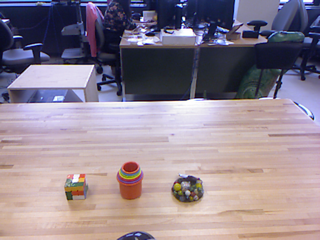}}}

\fcolorbox{white}{white}{\subfloat[Scene
17.  Seven objects: mustard, spam, soup, apple, jello, banana, pear.]{\label{scene_17}\includegraphics[width=0.24\textwidth]{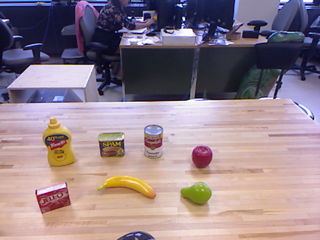}}}
\fcolorbox{white}{white}{\subfloat[Scene 18. Three objects: wood block, softball,  mug.]{\label{scene_18}\includegraphics[width=0.24\textwidth]{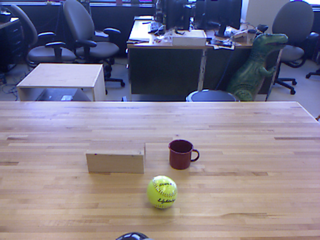}}}
\fcolorbox{white}{white}{\subfloat[Scene 19. Three objects: mustard, banana, pear.]{\label{scene_19}\includegraphics[width=0.24\textwidth]{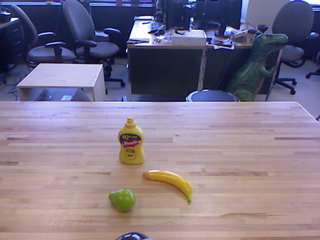}}}
\fcolorbox{white}{white}{\subfloat[Scene 20. Three objects: soccer ball, spam, jello.]{\label{scene_20}\includegraphics[width=0.24\textwidth]{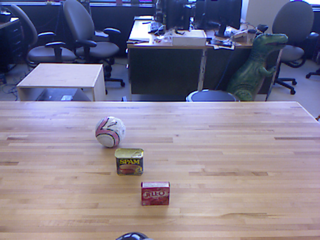}}}

  \caption{\label{scenes_pt1}Example test scenes (1-20)}
\end{figure*}

\begin{figure*}
  \centering
  
  \fcolorbox{white}{white}{\subfloat[Scene 21. Four objects: mug, softball, rubiks cube, mustard. The table is covered in a white tablecloth.]{\label{scene_21}\includegraphics[width=0.24\textwidth]{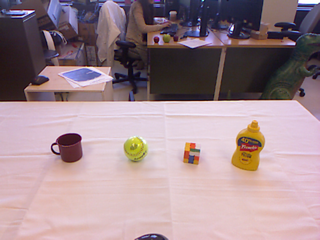}}}
  \fcolorbox{white}{white}{\subfloat[Scene
  22. Three objects: bowl, toy, soccer ball. The table is covered in a white tablecloth.]{\label{scene_22}\includegraphics[width=0.24\textwidth]{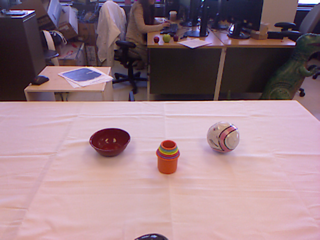}}}
  \fcolorbox{white}{white}{\subfloat[Scene 23.  Four objects: mustard, wood block, mug, soccer ball. The table is covered in a white tablecloth.]{\label{scene_23}\includegraphics[width=0.24\textwidth]{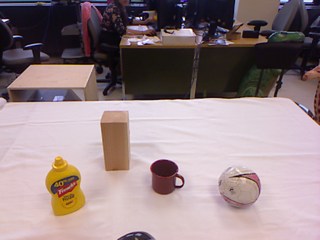}}}
  \fcolorbox{white}{white}{\subfloat[Scene 24. Five objects: ground coffee, cup, toy, bleach, softball. The table is covered in a white tablecloth.]{\label{scene_24}\includegraphics[width=0.24\textwidth]{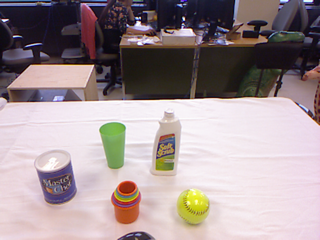}}}
  
  \fcolorbox{white}{white}{\subfloat[Scene 25. Four objects: mug, toy, soccer ball, banana. The table is covered in a white tablecloth.]{\label{scene_25}\includegraphics[width=0.24\textwidth]{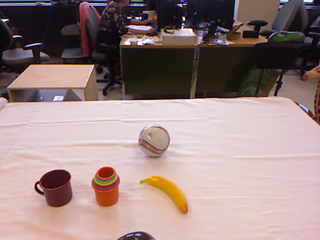}}}
  \fcolorbox{white}{white}{\subfloat[Scene 26. Three objects: mustard, ground coffee, softball.  The tabletop is dark brown.]{\label{scene_26}\includegraphics[width=0.24\textwidth]{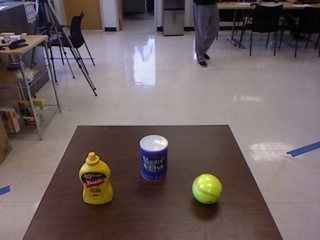}}}
  \fcolorbox{white}{white}{\subfloat[Scene
  27. Three objects: apple, soup, pear. The tabletop is dark brown.]{\label{scene_27}\includegraphics[width=0.24\textwidth]{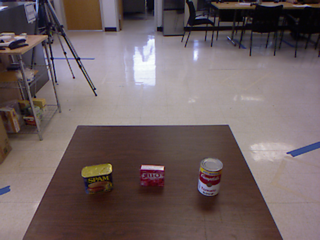}}}
  \fcolorbox{white}{white}{\subfloat[Scene 28.  Three objects: pringles, banana, cup. The tabletop is dark brown.]{\label{scene_28}\includegraphics[width=0.24\textwidth]{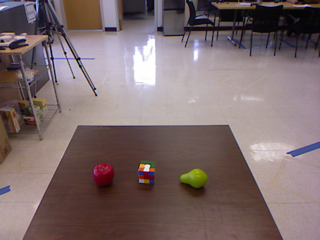}}}
  
  \fcolorbox{white}{white}{\subfloat[Scene 29. Three objects: pringles banana, cup. The tabletop is dark brown.]{\label{scene_29}\includegraphics[width=0.24\textwidth]{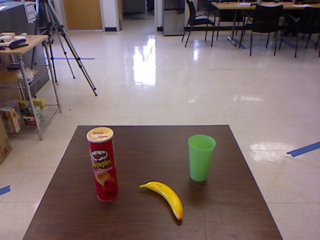}}}
  \fcolorbox{white}{white}{\subfloat[Scene 30.  Three objects: ground coffee, bowl, marbles. The tabletop is dark brown.]{\label{scene_30}\includegraphics[width=0.24\textwidth]{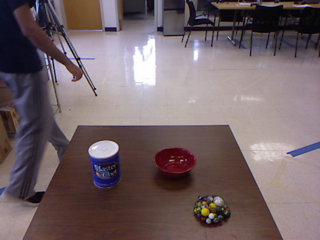}}}
  \fcolorbox{white}{white}{\subfloat[Scene 31. Three objects: mustard, softball, bowl. The camera was moving left to right.]{\label{scene_31}\includegraphics[width=0.24\textwidth]{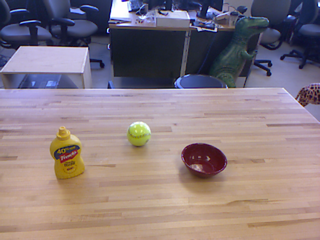}}}
  \fcolorbox{white}{white}{\subfloat[Scene 32. Three objects: wood block, pear, spam. The camera was moving left to right.]{\label{scene_32}\includegraphics[width=0.24\textwidth]{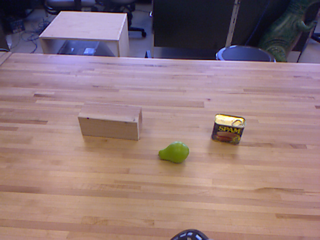}}}
  
  \fcolorbox{white}{white}{\subfloat[Scene 33.  Three objects: mug, marbles, soup. The camera was moving left to right.]{\label{scene_33}\includegraphics[width=0.24\textwidth]{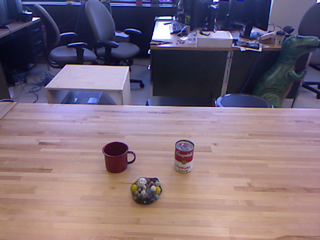}}}
  \fcolorbox{white}{white}{\subfloat[Scene 34. Three objects: apple, cup, jello. The camera was moving left to right.]{\label{scene_34}\includegraphics[width=0.24\textwidth]{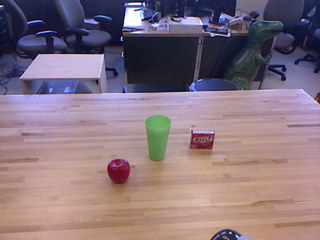}}}
  \fcolorbox{white}{white}{\subfloat[Scene 35. Three objects: pringles rubiks cube, banana. The camera was moving left to right.]{\label{scene_35}\includegraphics[width=0.24\textwidth]{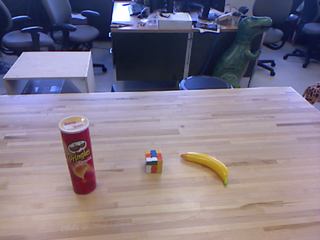}}}
  \fcolorbox{white}{white}{\subfloat[Scene 36.  Three objects: bleach, soccer ball, rubiks cube. The camera was moving from high to low.]{\label{scene_36}\includegraphics[width=0.24\textwidth]{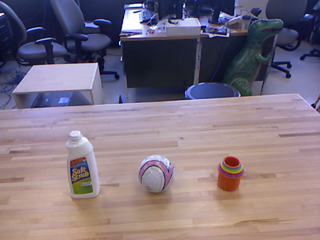}}}
  
  \fcolorbox{white}{white}{\subfloat[Scene 37. Three objects: mug, ground coffee, apple. The camera was moving from high to low.]{\label{scene_37}\includegraphics[width=0.24\textwidth]{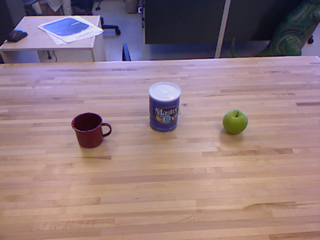}}}
  \fcolorbox{white}{white}{\subfloat[Scene 38. Three objects: cup, ground coffee, soccer ball. The camera was moving from high to low.]{\label{scene_38}\includegraphics[width=0.24\textwidth]{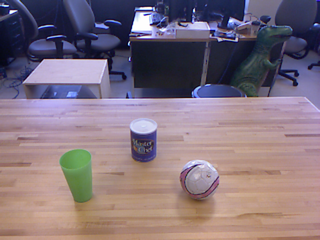}}}
  \fcolorbox{white}{white}{\subfloat[Scene 39. Three objects: mug, toy, jello. The camera was moving from high to low.]{\label{scene_39}\includegraphics[width=0.24\textwidth]{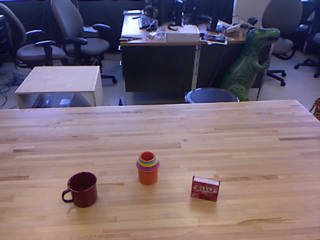}}}
  \fcolorbox{white}{white}{\subfloat[Scene 40. Three objects: mustard, softball, bowl. The camera was moving from high to low.]{\label{scene_40}\includegraphics[width=0.24\textwidth]{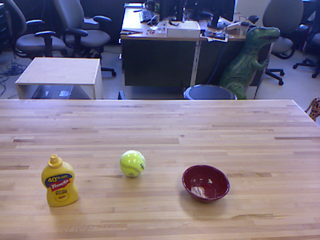}}}
  
  \caption{\label{scenes_pt2}Example test scenes (21-40)}
\end{figure*}

\end{document}